\pdfoutput=1
\documentclass[11pt]{article}
\usepackage[final]{acl}
\usepackage{times}
\usepackage{latexsym}
\usepackage[T1]{fontenc}
\usepackage[utf8]{inputenc}
\usepackage{microtype}
\usepackage{inconsolata}
\usepackage{dirtytalk}
\usepackage{placeins}
\usepackage{minitoc}
\usepackage{graphicx} 
\usepackage[utf8]{inputenc} 
\usepackage[T1]{fontenc}    
\usepackage{hyperref}       
\usepackage{url}            
\usepackage{booktabs}       
\usepackage{multirow}
\usepackage{amsfonts}       
\usepackage{nicefrac}       
\usepackage{microtype}      
\usepackage{xcolor}         
\usepackage{minitoc}
\usepackage{algorithm}
\usepackage{algpseudocode}
\usepackage{multicol}
\usepackage{tabularx}
\usepackage{pifont}
\usepackage{booktabs}
\usepackage{subcaption}
\usepackage{array}
\usepackage{makecell}

\newcommand{\xmark}{\ding{55}}

\usepackage[arrow, matrix, curve]{xy}
\usepackage{calc,multirow,tikz,array,setspace,geometry}

\usepackage{graphicx}
\usepackage{algorithmicx}
\usepackage{algpseudocode}
\usepackage{algorithm}
\usepackage{subcaption}
\usepackage{tikz}
\usetikzlibrary{calc,positioning}
\usetikzlibrary{matrix,calc,positioning,arrows}
\usetikzlibrary{arrows,shapes}
\usetikzlibrary{quotes,arrows.meta}
\usepackage{hyperref}
\usepackage{tcolorbox}
\usepackage{textcomp}
\usepackage{color}
\usepackage{enumitem}
\usepackage{wrapfig}
\usepackage{array}
\usepackage{amsthm}
\usepackage{amssymb}
\usepackage{amsmath}
\usepackage{bbold}

\theoremstyle{plain}

\theoremstyle{definition}

\theoremstyle{remark}

\newcommand{\vertiii}[1]{{\left\vert\kern-0.25ex\left\vert\kern-0.25ex\left\vert #1 
    \right\vert\kern-0.25ex\right\vert\kern-0.25ex\right\vert}}

\renewcommand{\geq}{\geqslant}

\usepackage{placeins}
\usepackage{bbm}
\usepackage[table]{xcolor}
\usepackage{booktabs}
\usepackage{makecell}
\usepackage{array}
\usepackage{siunitx}
\usepackage{xcolor}

\definecolor{charteOrange}{RGB}{255,121,0}
\definecolor{charteYellow}{RGB}{255,220,0}
\definecolor{charteViolet}{RGB}{145,100,205}
\definecolor{chartePink}{RGB}{255,180,230}
\definecolor{charteGreen}{RGB}{80,190,135}
\definecolor{charteBlue}{RGB}{75,180,230}
\definecolor{charteGray}{RGB}{143,143,143}

\usepackage{lipsum}

\newcommand\blfootnote[1]{%
  \begingroup
  \renewcommand\thefootnote{}\footnote{#1}%
  \addtocounter{footnote}{-1}%
  \endgroup
}

\title{SABET-QA: Temporal Knowledge Graph Question Answering}

\author{
\textbf{Brahim Touayouch\textsuperscript{1,2,*}}\quad
\textbf{Mirette Moawad\textsuperscript{1,*}}\quad
\textbf{Dmitry Akulov\textsuperscript{1,*}}
\\[0.6em]
\textsuperscript{1}QuickSort Research, Paris, France
\\
\textsuperscript{2}ENS Paris-Saclay, École Polytechnique, France
\\[0.4em]
\textbf{Contact:} \texttt{brahim.touayouch.2022@polytechnique.org}
}

\begin{document}

\tikzstyle{fleche} = [draw, thick, color=black, -latex']
\tikzstyle{dfleche} = [draw, thick, color=black, latex'-latex']
\tikzstyle{flecheDashed} = [draw, dashed, color=black, -latex']
\tikzstyle{DflecheDashed} = [draw, dashed, color=black, latex'-latex']

\newcommand{\st}{\,|\,}

\maketitle

\blfootnote{\textsuperscript{*} Work conducted at QuickSort Research, Paris, France (\url{https://www.quicksort.fr}). Correspondence to: {\tt mohamed@quicksort.fr} or personal emails.}

\begin{abstract}

    Question Answering over Temporal Knowledge Graphs (TKGQA) requires reasoning over time-sensitive facts, yet existing embedding-based methods struggle with multi-step queries due to single-pass reasoning pipelines. We propose SABET-QA, a framework that iteratively refines reasoning states across multiple hops via a bidirectional entity-temporal scoring mechanism and a slot-aware contextualization module that aligns question semantics with temporal KG embeddings. A differentiable working memory enables progressive hypothesis refinement, while auxiliary temporal boundaries serve as coarse supervision when available. Experiments on CronQuestions, Complex-CronQuestions, MultiTQ, and TimeQuestions demonstrate consistent improvements over strong baselines, particularly on complex multi-step temporal queries.
\end{abstract}

\section{Introduction}

\label{sec:introduction}

The proliferation of large-scale Knowledge Graphs (KGs) such as Wikidata~\cite{10.1145/2629489}, Freebase~\cite{10.1145/1376616.1376746} and YAGO~\cite{10.1145/1242572.1242667} has made Question Answering over KGs (KGQA) a crucial interface for accessing structured knowledge. However, real-world facts evolve over time, motivating Temporal Knowledge Graphs (TKGs) where a fact is represented as a quintuple $(s, r, o, [t_s, t_e])$. Temporal KGQA (TKGQA), the task of answering questions over such dynamic graphs, is essential for reasoning about a changing world.

Despite progress in static KGQA~\cite{jiang2023unikgqaunifiedretrievalreasoning, saxena-etal-2020-improving}, TKGQA presents unique challenges. Natural language questions contain explicit temporal constraints (e.g., ``Who was the president in 2008?'') or implicit compositional ones (e.g., ``Who was the president after Obama?''). Embedding-based approaches like CronKGQA~\cite{saxena-etal-2021-question} reduce simple queries to link prediction, but struggle with complex reasoning. Later work such as TempoQR~\cite{mavromatis2021tempoqrtemporalquestionreasoning} enriches question representations with contextualized time and entity information, yet these methods still process questions in a single forward pass, lacking iterative refinement and effective aggregation of distributed temporal evidence.

We identify three specific gaps in existing TKGQA methods:

\begin{enumerate}[nosep,topsep=0.3pt,leftmargin=*]
    \item \textbf{Single-Shot Reasoning:} Complex questions require sequential deduction (e.g., finding an entity, locating its tenure, then identifying the successor). Single-pass models cannot revisit or correct intermediate errors.
    \item \textbf{Ambiguous Directionality:} Questions mention entities without specifying their grammatical role (head or tail) in the relation. Existing models often assume a fixed direction, leading to incorrect scoring.
    \item \textbf{Context-Dependent Ambiguity:} Entities like ``Washington'' may refer to a person, a city, or a state depending on context. Lexical matching or fixed entity linking fails when the same mention carries different meanings.
\end{enumerate}

To address these gaps, we propose \textbf{SABET-QA}, a temporal KGQA framework combining slot-aware contextualization, bidirectional entity--temporal scoring, and iterative multi-hop reasoning. When coarse temporal hints are available, SABET-QA exploits them as additional supervision. A full architectural description is provided in Section~\ref{sec:imsabt}.

Our contributions are summarized as follows:

\begin{itemize}[nosep,topsep=0.3pt,leftmargin=*]
    \item We propose \textbf{SABET-QA}, an iterative temporal KGQA framework that progressively refines predictions through a working-memory-based multi-hop reasoning process.
    \item We introduce bidirectional entity--temporal scoring to address head--tail ambiguity and improve reasoning over questions with implicit directional structure.
    \item We empirically demonstrate that SABET-QA outperforms strong baselines on CronQuestions~\cite{saxena-etal-2021-question}, Complex-CronQuestions~\cite{CHEN2022109134}, MultiTQ~\cite{chen-etal-2023-multi}, and TimeQuestions~\cite{Jia_2021}, with particularly strong gains on complex questions.
\end{itemize}

\section{Related Work}

\label{sec:rel-work}

Our work relates to two areas: Temporal Knowledge Graph Representations and (Temporal) Knowledge Graph Question Answering.

\subsection{Temporal Knowledge Graph Representations}

Embedding-based TKGQA builds on Temporal Knowledge Graph Embedding (TKGE) methods. Early approaches such as TTransE~\cite{Leblay2018DerivingVT} extended static translation-based models~\cite{NIPS2013_1cecc7a7} by adding temporal embeddings to the scoring function. More recently, tensor decomposition methods have become prominent~\cite{Cai_2023}. TComplEx~\cite{lacroix2020tensordecompositionstemporalknowledge} extends ComplEx~\cite{trouillon2016complexembeddingssimplelink} to fourth-order tensors and shows that regularized decomposition can capture temporal dynamics effectively. These structured embedding spaces provide a compact latent representation of the TKG~\cite{cai2024surveytemporalknowledgegraph}, enabling downstream models to reason over relational and temporal dependencies in continuous vector form rather than through explicit graph traversal.

\subsection{Knowledge Graph Question Answering}

Knowledge graph question answering (KGQA) has evolved along three main directions: \emph{semantic parsing}, \emph{neural representation learning}, and \emph{LLM-based} approaches~\cite{su2026temporalknowledgegraphquestion}.

Semantic parsing methods~\cite{berant-etal-2013-semantic,chen2024selfimprovementprogrammingtemporalknowledge,yao-van-durme-2014-information,bao-etal-2016-constraint} translate natural-language questions into formal logical forms or executable structured queries. They offer explicit reasoning traces and high interpretability, but depend on handcrafted grammars, schema-specific operators, or substantial annotated supervision, limiting scalability on complex question types.

Neural representation learning approaches embed questions and graph elements into a shared latent space. Early work such as KEQA~\cite{10.1145/3289600.3290956} and EmbedKGQA~\cite{saxena-etal-2020-improving} treats QA as ranking over KG embeddings, avoiding explicit query construction. Later methods incorporate graph neural networks, attention mechanisms, and multi-hop reasoning to model structural dependencies and compositional questions~\cite{sun-etal-2018-open,Jia_2021,ijcai2023p571,10.1007/s10489-022-03913-6}. These models scale better than semantic parsers and tolerate noisy or incomplete graphs, but were developed for static KGs and do not model temporal validity.

LLM-based methods extend KGQA beyond fixed templates by generating executable queries, reasoning over retrieved evidence, or combining retrieval with generation~\cite{qian-etal-2024-timer4,jia2024faithfultemporalquestionanswering,gao2024twostagegenerativequestionanswering,chen-etal-2024-temporal}. They reduce hand-engineered logic and improve linguistic flexibility, yet remain sensitive to retrieval quality, grounding errors, and hallucination, and are still most mature in non-temporal settings.

These limitations are amplified in temporal KGQA, where answers depend on when facts hold. The field has therefore moved from direct retrieval toward multi-step reasoning over time-sensitive constraints. CronKGQA~\cite{saxena-etal-2021-question} casts a question as a \emph{virtual relation} in a temporal embedding space, enabling link-prediction-style answering. This works for simple questions but struggles with sequential deduction or head--tail ambiguity. TempoQR~\cite{mavromatis2021tempoqrtemporalquestionreasoning} enriches question representations with contextualized temporal and entity information, yielding stronger performance on complex questions, but its reasoning remains largely single-pass, limiting the ability to revise intermediate hypotheses. SubGTR~\cite{CHEN2022109134} introduces subgraph-based temporal reasoning with logical constraints and highlights pseudo-temporal questions in CronQuestions. While effective when subgraph extraction is reliable, it depends heavily on extracted structures and is less robust on incomplete or noisy graphs, or when transferring across datasets.

In contrast, SABET-QA follows the embedding-based virtual-relation paradigm but adds iterative multi-hop reasoning with differentiable working memory. At each hop, the model refines a latent reasoning state, builds hop-specific relation representations, and computes intermediate scores for both entities and timestamps. To address head--tail ambiguity, it uses bidirectional entity scoring, combining forward and backward signals through a learned gate. Unlike approaches relying on explicit subgraph extraction or dataset-specific post-processing, SABET-QA operates directly over pretrained temporal KG embeddings and natural-language questions, making it broadly applicable across benchmarks while remaining simple in design.

Recent work also explores LLM-based autonomous agents for TKGQA, but these introduce substantial inference latency, computational overhead, and dependency on non-deterministic external APIs. SABET-QA operates strictly within the dense embedding-based paradigm, providing low-latency, deterministic, and locally deployable inference. We therefore compare against embedding-based baselines as the appropriate reference class.

\section{Temporal Knowledge Graph Question Answering Methods}
\subsection{Temporal KG Embeddings}

We train the temporal knowledge graph embeddings used throughout our models with \textit{TComplEx}~\cite{lacroix2020tensordecompositionstemporalknowledge}. TComplEx represents entities, relations, and timestamps with complex-valued embeddings and scores a temporal fact $(s,r,o,t)$ using a multilinear complex product:
\[
\phi(s,r,o,t) = \Re\left(\langle \mathbf{u}_s, \mathbf{v}_r, \overline{\mathbf{u}}_o, \mathbf{w}_t \rangle\right),
\]
where $\mathbf{u}_s$, $\mathbf{u}_o$, $\mathbf{v}_r$, and $\mathbf{w}_t$ denote the subject, object, relation, and timestamp embeddings, respectively.


The learned entity and time representationss are used to initialize the shared embedding tables across all QA models \footnote{The embeddings used for the datasets in this study are provided in the supplementary material. To facilitate future work, we standardized the embedding structure across all datasets. Some embeddings were trained from scratch, whereas others were initialized from prior work.}. Additional implementation and training details are provided in Appendix~\ref{app:tcomplex}.

\subsection{Baseline Methods}

This section summarizes the main comparison baselines used in our experiments.

\subsubsection{LM\_TKGQA: Language Model Baseline}

LM\_TKGQA encodes the question with a pretrained language model such as BERT~\cite{devlin2019bertpretrainingdeepbidirectional} or RoBERTa~\cite{liu2019robertarobustlyoptimizedbert} and projects it into the temporal knowledge graph embedding space. Candidate entities and timestamps are then ranked with a lightweight prediction head, but temporal structure is not modeled explicitly.

\subsubsection{EmbedKGQA: Embedding-Based KGQA}

EmbedKGQA represents the question as a soft relation and retrieves answers by matching it against knowledge graph embeddings~\cite{saxena-etal-2020-improving}. While effective for multi-hop reasoning, it does not explicitly model temporal constraints.

\subsubsection{CRONKGQA: Temporal Inference Model}

CRONKGQA extends embedding-based KGQA by jointly predicting entities and timestamps using temporal knowledge graph embeddings \cite{saxena-etal-2021-question}. This enables reasoning over temporal ordering and time-sensitive facts.

\subsubsection{TempoQR: Contextualized Temporal Reasoning}

TempoQR combines a pretrained language model, temporal knowledge graph embeddings, and a transformer-based fusion module \cite{mavromatis2021tempoqrtemporalquestionreasoning}. This improves contextual understanding while explicitly incorporating temporal information.

\subsubsection{SubGTR: Subgraph-Based Temporal Reasoning}

SubGTR~\cite{CHEN2022109134} performs reasoning over a task-relevant subgraph extracted around the query and its temporal context. Restricting inference to this neighborhood improves efficiency while exploiting local graph structure.

\paragraph{On postprocessing and dataset dependence.}
Some baselines, such as SubGTR, rely on additional postprocessing (e.g., subgraph extraction) tailored to specific datasets. In contrast, our model is designed to operate across datasets without task-specific processing, making it easier to transfer between benchmarks.

\subsection{Proposed Method: SABET-QA}
\label{sec:imsabt}

\begin{figure*}[t]
    \centering
    \includegraphics[width=0.98\columnwidth]{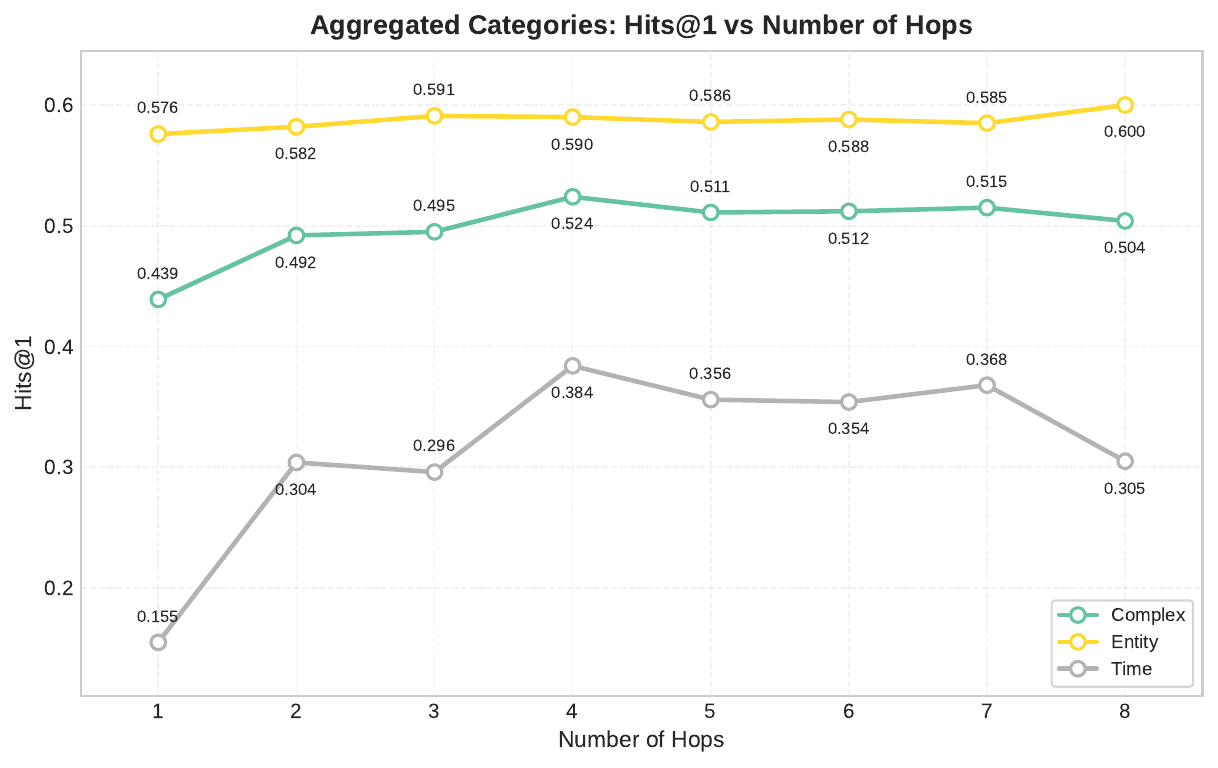}
    \includegraphics[width=0.98\columnwidth]{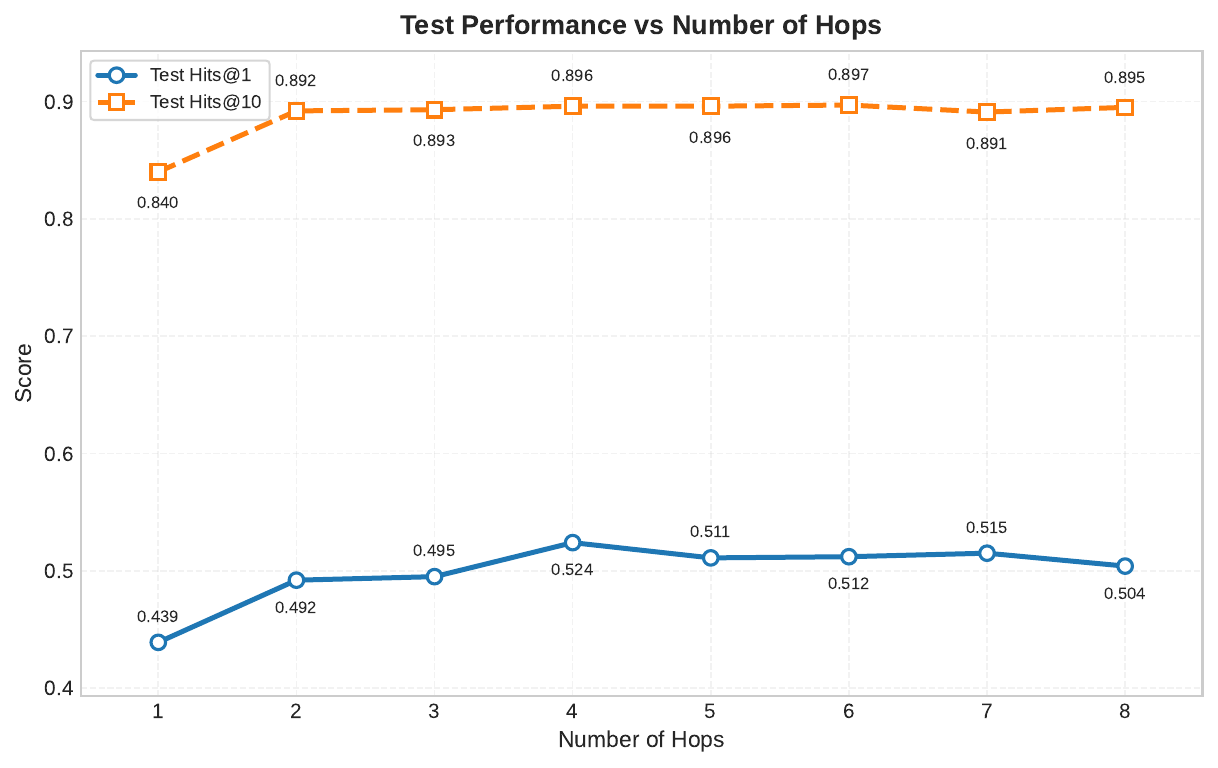}
    \caption{Impact of the number of reasoning hops on Complex-CronQuestions. Left: Hits@1 for different question categories. Right: overall test Hits@1 and Hits@10.}
    \label{fig:hops_analysis}
\end{figure*}

\begin{table*}[tb]
\centering
\small
\setlength{\tabcolsep}{5pt}
\renewcommand{\arraystretch}{1.15}
\begin{tabular}{lccc|cc}
\toprule
\textbf{Model} &
\textbf{Entity Contextualization} &
\textbf{\makecell{Bidirectional Entity\\\& Time Scoring}} &
\textbf{Hard Supervision} &
\multicolumn{2}{c}{\textbf{Complex-CronQuestions}} \\
\cmidrule(lr){5-6}
& & & & \textbf{Hits@1} & \textbf{Hits@10} \\
\midrule

\multirow{4}{*}{\textbf{SABET-QA-Hard}}
& \cellcolor{gray!15}\checkmark
& \cellcolor{gray!15}\checkmark
& \cellcolor{gray!15}\checkmark
& \cellcolor{gray!15}\textbf{0.807}
& \cellcolor{gray!15}\textbf{0.962} \\
& \xmark     & \checkmark & \checkmark & 0.707 & 0.946 \\
& \checkmark & \xmark     & \checkmark & 0.659 & 0.849 \\
& \checkmark & \checkmark & \xmark     & 0.524 & 0.896 \\
\bottomrule
\end{tabular}
\caption{
Ablation study of SABET-QA-Hard on the Complex-CronQuestions test set. Removing any component degrades performance, with the largest drop observed when bidirectional entity and time scoring is disabled.
}
\label{tab:ablation}
\end{table*}

\begin{table*}[tb]
\centering
\small
\setlength{\tabcolsep}{5pt}
\renewcommand{\arraystretch}{1.15}
\begin{tabular}{lcc|cc|cc}
\toprule
\textbf{Model} &
\textbf{Frozen LM} &
\textbf{Frozen TKE} &
\multicolumn{2}{c|}{\textbf{Complex-CronQuestions}} &
\multicolumn{2}{c}{\textbf{CronQuestions}} \\
\cmidrule(lr){4-5} \cmidrule(lr){6-7}
& & & \textbf{Hits@1} & \textbf{Hits@10} & \textbf{Hits@1} & \textbf{Hits@10} \\
\midrule

\multirow{4}{*}{\textbf{SABET-QA}}
& \cellcolor{gray!15}\checkmark
& \cellcolor{gray!15}\checkmark
& \cellcolor{gray!15}0.524
& \cellcolor{gray!15}\textbf{0.896}
& \cellcolor{gray!15}\textbf{0.843}
& \cellcolor{gray!15}\textbf{0.969} \\
& \checkmark & \xmark     & 0.446 & 0.871 & 0.773 & 0.950 \\
& \xmark     & \checkmark & \textbf{0.547} & 0.892 & 0.836 & 0.968 \\
& \xmark     & \xmark     & 0.476 & 0.882 & 0.784 & 0.946 \\
\midrule

\multirow{4}{*}{\textbf{SABET-QA-Hard}}
& \cellcolor{gray!15}\checkmark
& \cellcolor{gray!15}\checkmark
& \cellcolor{gray!15}\textbf{0.807}
& \cellcolor{gray!15}0.962
& \cellcolor{gray!15}\textbf{0.954}
& \cellcolor{gray!15}\textbf{0.989} \\
& \checkmark & \xmark     & 0.759 & 0.958 & 0.925 & 0.983 \\
& \xmark     & \checkmark & 0.803 & \textbf{0.964} & 0.949 & 0.988 \\
& \xmark     & \xmark     & 0.769 & 0.959 & 0.902 & 0.974 \\
\bottomrule
\end{tabular}
\caption{
Ablation study evaluating the impact of freezing the language model (LM) and temporal knowledge graph embedding (TKE) modules under both no supervision (\textbf{SABET-QA}) and hard supervision (\textbf{SABET-QA-Hard}). The comparison isolates the contribution of each component and demonstrates the effect of unfreezing the parameters on overall reasoning performance.
}
\label{tab:ablation_freeze}
\end{table*}

SABET-QA answers temporal questions through \textbf{iterative latent-state refinement} over $K$ hops. At each hop, the model produces bidirectional entity and time scores, then updates a differentiable working memory to progressively sharpen its hypothesis. The architecture couples a frozen pretrained LM with a TComplEx temporal scorer, operating in four stages: (i) question encoding and slot contextualization, (ii) hop-specific relation projection, (iii) bidirectional scoring with memory update, and (iv) adaptive hop aggregation. Pseudocode and full dimensional details are in Appendix~\ref{app:imsabt}.

\subsubsection{Slot-Aware Question Encoding}

Given tokenized question $\mathbf{x}$, a pretrained LM yields hidden states $\mathbf{H} \in \mathbb{R}^{L \times 768}$, where $L$ is the number of tokens and $768$ is the hidden representation dimension of the pretrained LM (BERT or RoBERTa). These representations are linearly projected to the TKG embedding dimension:

\begin{equation}
\mathbf{T} = f_{\text{text}}(\mathbf{H}) \in \mathbb{R}^{L \times D}.
\end{equation} where $D$ denotes the dimensionality of the TKG embedding space (i.e., the entity and timestamp embedding dimension).

The model first identifies the entity and temporal mentions in the question. When gold annotations are available, these mentions are extracted directly; otherwise, they are obtained using a named entity recognition (NER) system. The identified entity mentions are arbitrarily assigned to the head ($\mathbf{h}$) and tail ($\mathbf{t}$) query slots, while the temporal expression is assigned to the time slot ($\boldsymbol{\tau}$). The corresponding embeddings are retrieved from the TKBC embedding tables, with learned dummy embeddings used for any missing slots. These query embeddings are then contextualized via multi-head cross-attention~\cite{vaswani2023attentionneed} over the projected question token representations, followed by a residual gating mechanism:
\begin{equation}
[\mathbf{h}', \mathbf{t}', \boldsymbol{\tau}'] =
\mathrm{Gate}\!\left(
\mathrm{MHA}([\mathbf{h}, \mathbf{t}, \boldsymbol{\tau}], \mathbf{T}, \mathbf{T})
\right).
\end{equation}


A global question representation $\mathbf{s}$ is constructed by concatenating the hidden representation of the special classification token ([CLS]), the mean-pooled token representations, and the max-pooled token representations of $\mathbf{T}$. The resulting vector is projected to obtain the base reasoning state:
\begin{equation}
\mathbf{z}^{(0)} = f_{\text{rel}}\big([\mathbf{s}; \mathbf{h}'; \mathbf{t}'; \boldsymbol{\tau}']\big).
\end{equation}

\subsubsection{Iterative Hop-Wise Reasoning with Bidirectional Scoring}

At hop $k$, the current latent state $\mathbf{z}^{(k-1)}$ and summary $\mathbf{s}$ are fused into a hop-specific relation vector $\mathbf{r}^{(k)} = f^{(k)}_{\text{hop}}([\mathbf{z}^{(k-1)}; \mathbf{s}])$, then split into entity-oriented and time-oriented projections: $\mathbf{r}^{(k)}_{\text{ent}}$ and $\mathbf{r}^{(k)}_{\text{time}}$.

\paragraph{Temporal hint injection (optional).}
When auxiliary temporal boundaries $(t_1, t_2)$ are available, we provide them as hard supervision (denoted by the \textit{hard} suffix in the model name). Following TempoQR~\cite{mavromatis2021tempoqrtemporalquestionreasoning}, these boundaries are derived from the earliest and latest timestamps of facts involving the question entities and are injected into the working memory through a learned gating mechanism:
\begin{equation}
\mathbf{z}^{(k-1)} \leftarrow \mathbf{z}^{(k-1)} + \boldsymbol{\gamma}^{(k)}_{\text{time}} \odot (\mathbf{e}_{t_1} + \mathbf{e}_{t_2}).
\end{equation}

\paragraph{Bidirectional scoring.}
To resolve head-tail ambiguity, entity candidates are scored in both directions using the contextualized slots, then fused by a summary-conditioned gate $\alpha^{(k)}_{\text{ent}} = \sigma(W_{\text{ent}}\mathbf{s})$:


\vspace{-1pt}
{
\setlength{\abovedisplayskip}{2pt}
\setlength{\abovedisplayshortskip}{2pt}
\setlength{\belowdisplayskip}{2pt}
\setlength{\belowdisplayshortskip}{2pt}

\centering
\resizebox{\linewidth}{!}{%
\begin{minipage}{\linewidth}
\begin{multline}
\mathbf{s}^{(k)}_{\text{ent}} =
\alpha^{(k)}_{\text{ent}} \cdot
\text{Score}_{\text{TComplEx}}(\mathbf{h}', \mathbf{t}', \mathbf{r}^{(k)}_{\text{ent}}, \boldsymbol{\tau}') \\
+ (1-\alpha^{(k)}_{\text{ent}}) \cdot
\text{Score}_{\text{TComplEx}}(\mathbf{t}', \mathbf{h}', \mathbf{r}^{(k)}_{\text{ent}}, \boldsymbol{\tau}')
\end{multline}
\end{minipage}%
}
\par
}

\vspace{3pt}
Timestamp candidates are scored analogously (without the time slot in the scorer) and fused via $\alpha^{(k)}_{\text{time}}$.

\subsubsection{Working-Memory Update and Aggregation}

After scoring, soft distributions $\mathbf{p}^{(k)}_{\text{ent}} = \text{softmax}(\mathbf{s}^{(k)}_{\text{ent}})$ and $\mathbf{p}^{(k)}_{\text{time}} = \text{softmax}(\mathbf{s}^{(k)}_{\text{time}})$ yield expected embeddings:
\begin{equation}
\bar{\mathbf{e}}^{(k)}_{\text{ent}} = \mathbf{p}^{(k)\top}_{\text{ent}}\mathbf{E}_{\text{ent}}, \quad
\bar{\mathbf{e}}^{(k)}_{\text{time}} = \mathbf{p}^{(k)\top}_{\text{time}}\mathbf{E}_{\text{time}}.
\end{equation}

Their sum is projected to a memory vector $\mathbf{m}^{(k)}$ that refines the latent state:
\begin{equation}
\mathbf{z}^{(k)} = \text{Gate}\big(\text{Attn}(\mathbf{z}^{(k-1)}, \mathbf{m}^{(k)})\big).
\end{equation}

This lets the model carry forward soft predictions from earlier hops, progressively refining its hypothesis.

\paragraph{Adaptive aggregation.}
A hop selector $\boldsymbol{\beta} = \text{softmax}(W_{\text{hop}}\mathbf{s})$ weights the $K$ hop-specific score vectors:
\begin{equation}
\mathbf{s}_{\text{ent}} = \sum_{k=1}^{K} \beta_k \mathbf{s}^{(k)}_{\text{ent}}, \qquad
\mathbf{s}_{\text{time}} = \sum_{k=1}^{K} \beta_k \mathbf{s}^{(k)}_{\text{time}}.
\end{equation}

The concatenated output $[\mathbf{s}_{\text{ent}}; \mathbf{s}_{\text{time}}]$ is trained with cross-entropy against the gold answer. TKBC embeddings remain fixed during QA training unless the unfrozen ablation setting is enabled.

\section{Experiments}
\label{sec:expe}

\subsection{Datasets}

We evaluate SABET-QA on four benchmark datasets for temporal question answering: CronQuestions~\cite{saxena-etal-2021-question}, Complex-CronQuestions~\cite{CHEN2022109134}, MultiTQ~\cite{chen-etal-2023-multi}, and TimeQuestions~\cite{Jia_2021}. Together, these benchmarks cover a broad spectrum of temporal reasoning settings, including temporal entity prediction, timestamp prediction, temporal comparison, and multi-hop reasoning over temporal knowledge graphs.

CronQuestions and Complex-CronQuestions are synthetic benchmarks derived from temporal Wikidata facts and are designed to assess compositional temporal reasoning. MultiTQ is a large-scale automatically generated dataset containing questions that involve diverse temporal operators and more intricate reasoning patterns. TimeQuestions comprises natural-language temporal questions collected from real-world sources, offering a complementary evaluation setting that more closely reflects realistic user queries.

Detailed dataset statistics, temporal knowledge graph characteristics, and answer-type distributions are provided in Appendix~\ref{app:datasets} (Tables~\ref{tab:dataset_meta}, \ref{tab:tkg_meta}, and \ref{tab:answer_type_distributions}).

\subsection{Method Evaluation}

\begin{table*}[tb]
\centering
\small
\begin{tabular}{lcccccccccc}
\toprule
\multirow{2}{*}{Model} & \multicolumn{5}{c}{Hits@1} & \multicolumn{5}{c}{Hits@10} \\
\cmidrule(lr){2-6} \cmidrule(lr){7-11}
& Overall & Complex & Simple & Entity & Time & Overall & Complex & Simple & Entity & Time \\
\midrule
BERT              & 0.257 & 0.253 & 0.262 & 0.292 & 0.191 & 0.642 & 0.617 & 0.676 & 0.650 & 0.627 \\
CronKGQA          & 0.646 & 0.391 & 0.987 & 0.698 & 0.550 & 0.886 & 0.806 & 0.993 & 0.901 & 0.858 \\
EmbedKGQA         & 0.433 & 0.370 & 0.516 & 0.576 & 0.166 & 0.787 & 0.735 & 0.857 & 0.892 & 0.592 \\
TempoQR           & 0.796 & 0.658 & 0.981 & 0.880 & 0.640 & 0.959 & 0.934 & 0.992 & 0.975 & 0.930 \\
TempoQR-Hard      & 0.914 & 0.861 & 0.984 & 0.923 & 0.896 & 0.979 & 0.968 & 0.993 & 0.982 & 0.973 \\
SubGTR-Hard       & 0.913 & 0.859 & 0.984 & 0.917 & 0.904 & 0.980 & 0.970 & 0.993 & 0.982 & 0.975 \\
\midrule
SABET-QA          & \textbf{0.843} & \textbf{0.733} & \textbf{0.989} & \textbf{0.882} & \textbf{0.770} & \textbf{0.969} & \textbf{0.953} & \textbf{0.994} & \textbf{0.979} & \textbf{0.954} \\
SABET-QA-Hard     & \textbf{0.954} & \textbf{0.926} & \textbf{0.994} & \textbf{0.941} & \textbf{0.980} & \textbf{0.989} & \textbf{0.983} & \textbf{0.996} & \textbf{0.986} & \textbf{0.994} \\
\bottomrule
\end{tabular}
\caption{Comparison against other methods on the CronQuestions test set. Metrics are reported for overall performance, question type (complex/simple), and answer type (entity/time).}
\label{tab:cronquestions_comparison}
\end{table*}

\begin{table*}[tb]
\centering
\small
\begin{tabular}{lcccccc}
\toprule
\multirow{2}{*}{Model} & \multicolumn{3}{c}{Hits@1} & \multicolumn{3}{c}{Hits@10} \\
\cmidrule(lr){2-4} \cmidrule(lr){5-7}
& Overall & Entity & Time & Overall & Entity & Time \\
\midrule
BERT              & 0.087 & 0.097 & 0.068 & 0.421 & 0.351 & 0.567 \\
CronKGQA          & 0.288 & 0.365 & 0.129 & 0.736 & 0.758 & 0.689 \\
EmbedKGQA         & 0.260 & 0.361 & 0.050 & 0.618 & 0.742 & 0.360 \\
TempoQR           & 0.438 & 0.585 & 0.132 & 0.853 & 0.906 & 0.743 \\
TempoQR-Hard      & 0.632 & 0.721 & 0.448 & 0.933 & 0.942 & 0.914 \\
SubGTR-Hard       & 0.623 & 0.719 & 0.422 & 0.928 & 0.944 & 0.895 \\
\midrule
SABET-QA          & 0.524 & 0.590 & 0.384 & 0.896 & 0.925 & 0.836 \\
SABET-QA-Hard     & \textbf{0.807} & \textbf{0.747} & \textbf{0.931} & \textbf{0.962} & \textbf{0.955} & \textbf{0.978} \\
\bottomrule
\end{tabular}

\caption{Comparison on the Complex-CronQuestions test set. Metrics are reported for overall performance and answer type (entity/time).}
\label{tab:complex_cronquestions_comparison}
\end{table*}

\begin{table*}[tb]
\centering
\small
\setlength{\tabcolsep}{4pt}
\renewcommand{\arraystretch}{1.15}
\resizebox{\textwidth}{!}{%
\begin{tabular}{lcccccc|cccccc}
\toprule
\multirow{3}{*}{\textbf{Model}} &
\multicolumn{6}{c|}{\textbf{MultiTQ}} &
\multicolumn{6}{c}{\textbf{TimeQuestions}} \\
\cmidrule(lr){2-7} \cmidrule(lr){8-13}
& \multicolumn{3}{c}{\textbf{Hits@1}} &
\multicolumn{3}{c|}{\textbf{Hits@10}} &
\multicolumn{3}{c}{\textbf{Hits@1}} &
\multicolumn{3}{c}{\textbf{Hits@10}} \\
\cmidrule(lr){2-4}
\cmidrule(lr){5-7}
\cmidrule(lr){8-10}
\cmidrule(lr){11-13}
& \textbf{Overall} & \textbf{Entity} & \textbf{Time}
& \textbf{Overall} & \textbf{Entity} & \textbf{Time}
& \textbf{Overall} & \textbf{Entity} & \textbf{Time}
& \textbf{Overall} & \textbf{Entity} & \textbf{Time} \\
\midrule
BERT         & 0.103 & 0.117 & 0.069 & 0.516 & 0.632 & 0.234 & 0.450 & 0.409 & 0.556 & 0.569 & 0.520 & 0.698 \\
CronKGQA     & 0.278 & 0.387 & 0.011 & 0.527 & 0.724 & 0.046 & 0.326 & 0.296 & 0.405 & 0.454 & 0.409 & 0.569 \\
EmbedKGQA    & 0.243 & 0.342 & 0.002 & 0.489 & 0.685 & 0.012 & 0.288 & 0.266 & 0.344 & 0.468 & 0.426 & 0.577 \\
TempoQR      & 0.327 & 0.456 & 0.014 & 0.571 & 0.783 & 0.055 & 0.409 & 0.406 & 0.416 & 0.530 & 0.513 & 0.574 \\
TempoQR-Hard & 0.335 & 0.465 & 0.018 & 0.579 & 0.788 & 0.068 & 0.410 & 0.411 & 0.407 & 0.528 & 0.511 & 0.572 \\
SubGTR-Hard  & 0.337 & 0.469 & 0.015 & 0.576 & 0.789 & 0.056 & 0.419 & 0.415 & 0.427 & 0.532 & 0.512 & 0.583 \\
\midrule
SABET-QA      & 0.373 & \textbf{0.480} & 0.111 & 0.700 & 0.804 & 0.444 & 0.502 & 0.500 & \textbf{0.609} & \textbf{0.619} & \textbf{0.568} & \textbf{0.750} \\
SABET-QA-Hard & \textbf{0.403} & 0.479 & \textbf{0.219} & \textbf{0.715} & \textbf{0.810} & \textbf{0.485} & \textbf{0.504} & \textbf{0.513} & 0.604 & 0.615 & 0.565 & 0.747 \\
\bottomrule
\end{tabular}%
}
\caption{Comparison on the MultiTQ and TimeQuestions test sets.}
\label{tab:multitq_timequestions_comparison}
\end{table*}

We evaluate SABET-QA using the standard ranking-based metrics commonly adopted in the temporal question answering literature, namely \textit{Hits@1} and \textit{Hits@10}. Hits@1 measures the proportion of queries for which the correct answer is ranked first, whereas Hits@10 measures the proportion of queries for which the correct answer appears within the top ten ranked candidates. Following the evaluation protocols of the respective benchmark datasets, we report overall performance as well as disaggregated results by question complexity (when available) and answer type (entity versus timestamp).

Table~\ref{tab:cronquestions_comparison} reports results on CronQuestions.
SABET-QA achieves \textbf{0.843} Hits@1, improving over TempoQR by
\textbf{4.7} points, a gain concentrated almost entirely on the complex
subset (\textbf{+7.5} over TempoQR's 0.658).  This pattern, strong
gains on compositional reasoning with maintained or improved performance
on simple queries, recurs across all four benchmarks and validates our
core hypothesis: iterative refinement with bidirectional scoring is
particularly effective when questions require sequential temporal
deduction.

The \emph{hard-supervised} variant (SABET-QA-Hard) pushes performance
to \textbf{0.954} Hits@1 on CronQuestions and \textbf{0.807} on
Complex-CronQuestions (Table~\ref{tab:complex_cronquestions_comparison}).
The margin over TempoQR-Hard widens to \textbf{17.5} points on
Complex-CronQuestions, suggesting that our architecture better exploits
coarse temporal boundaries when reasoning chains are longer.  Notably,
SABET-QA-Hard's time-answer accuracy on Complex-CronQuestions
(\textbf{0.931} Hits@1) approaches its entity-answer accuracy
(\textbf{0.747}), whereas TempoQR-Hard exhibits a \textbf{27.3}-point
gap between the two.  We attribute this to the working-memory mechanism,
which propagates intermediate temporal hypotheses across hops rather
than scoring time and entity candidates independently.

On MultiTQ (Table~\ref{tab:multitq_timequestions_comparison}), absolute scores are
lower across all methods, reflecting the dataset's coarser temporal
granularity and more diverse operator vocabulary.  SABET-QA still leads
all baselines, with the largest relative improvement on timestamp
prediction (\textbf{0.111} vs.\ Bert's \textbf{0.069} Hits@1).

TimeQuestions (Table~\ref{tab:multitq_timequestions_comparison}) presents
naturalistic questions with noisy entity linking.  Here SABET-QA
outperforms TempoQR by \textbf{9.3} Hits@1 points overall, with the
non-hard variant actually edging SABET-QA-Hard on time-specific
accuracy (\textbf{0.609} vs.\ \textbf{0.604}).

\paragraph{Overall,} SABET-QA establishes the best results on all four benchmarks.  The gains are
largest on complex multi-hop questions and timestamp prediction,
confirming that iterative, structure-aware temporal reasoning
outperforms single-pass embedding matching.

\subsection{Method Behaviour Analysis}

\paragraph{Varying the number of hops.}
Figure~\ref{fig:hops_analysis} (left) plots per-category Hits@1 on
Complex-CronQuestions as $K$ increases from 1 to 8.  Three distinct
regimes emerge.  \emph{Complex} queries rise sharply from $K{=}1$
(\textbf{0.439}) to $K{=}4$ (\textbf{0.524}), then fluctuate in a
narrow band (\textbf{0.504} - \textbf{0.515}) through $K{=}8$;
the peak at $K{=}4$ represents a \textbf{19.4}\% relative gain over
single-hop reasoning.  \emph{Time} queries follow a similar trajectory,
peaking at $K{=}4$ (\textbf{0.384}) before degrading to \textbf{0.305}
at $K{=}8$.

\emph{Entity} queries behave differently: they rise
monotonically from $K{=}1$ (\textbf{0.576}) to $K{=}8$
(\textbf{0.600}), with only marginal gains beyond $K{=}4$.
This divergence is structurally informative: complex and time queries
benefit from \emph{moderate} iterative refinement enough to resolve
intermediate ambiguities, but not so deep that compounding complexity
dominates whereas entity queries, which require less compositional
reasoning, tolerate and even profit from extended unrolling.

The right panel confirms that overall Hits@1 mirrors the complex
category, peaking at $K{=}4$ (\textbf{0.524}).  Hits@10 behaves
differently: it rises steeply from $K{=}1$ (\textbf{0.840}) to
$K{=}2$ (\textbf{0.892}), then plateaus from $K{=}4$ onward
(\textbf{0.895} --- \textbf{0.897}).  This decoupling Hits@1
saturating earlier than Hits@10 indicates that additional hops
primarily improve ranking quality within the top 10 rather than
top-1 precision.  We use $K{=}4$ in all reported results as the
best compromise between accuracy and computational cost.

\begin{figure*}[t!]
    \centering
    \begin{subfigure}{0.32\textwidth}
        \includegraphics[width=\linewidth]{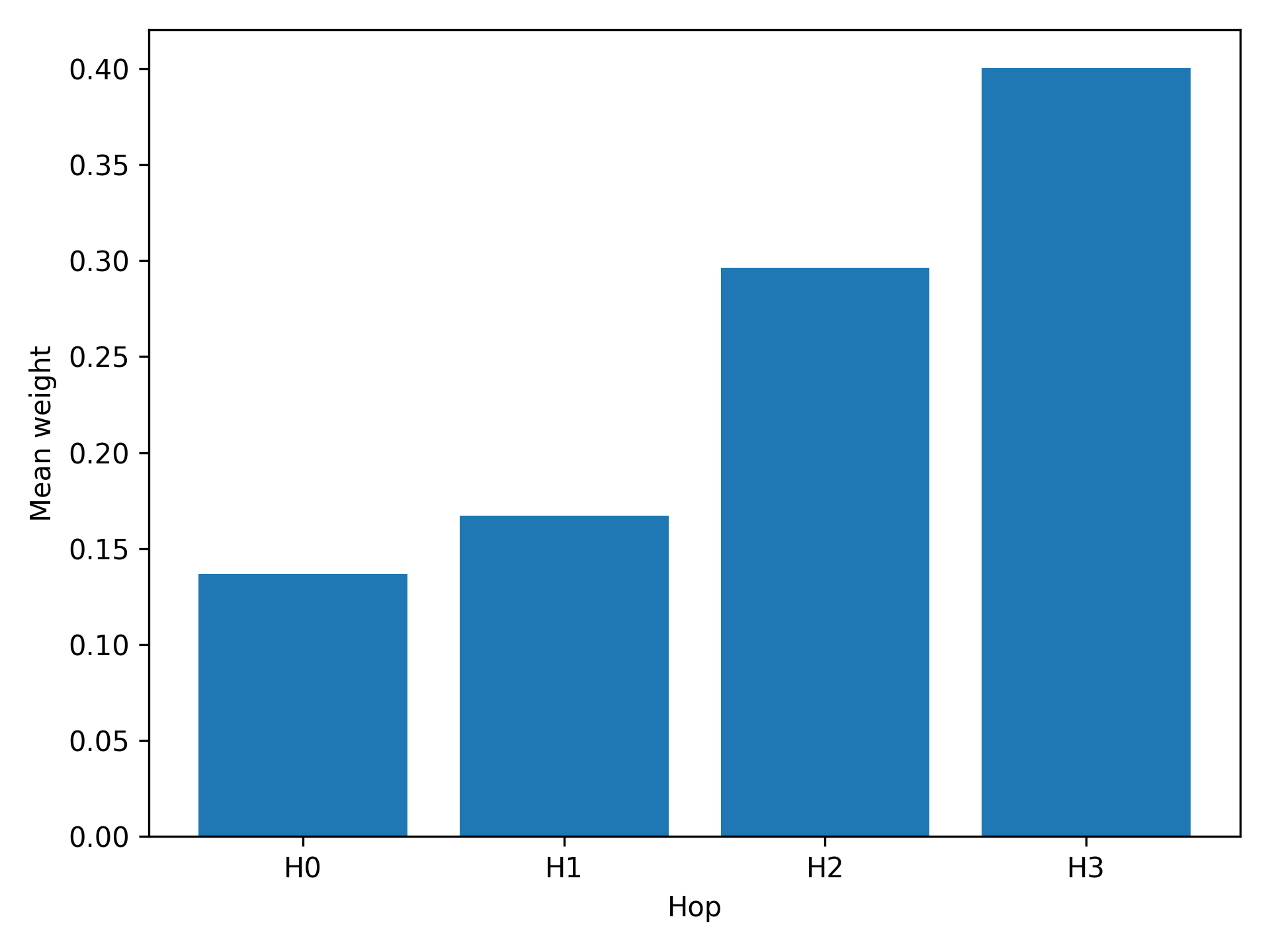}
        \caption{Mean aggregation weight ($\beta_k$).}
    \end{subfigure}\hfill
    \begin{subfigure}{0.32\textwidth}
        \includegraphics[width=\linewidth]{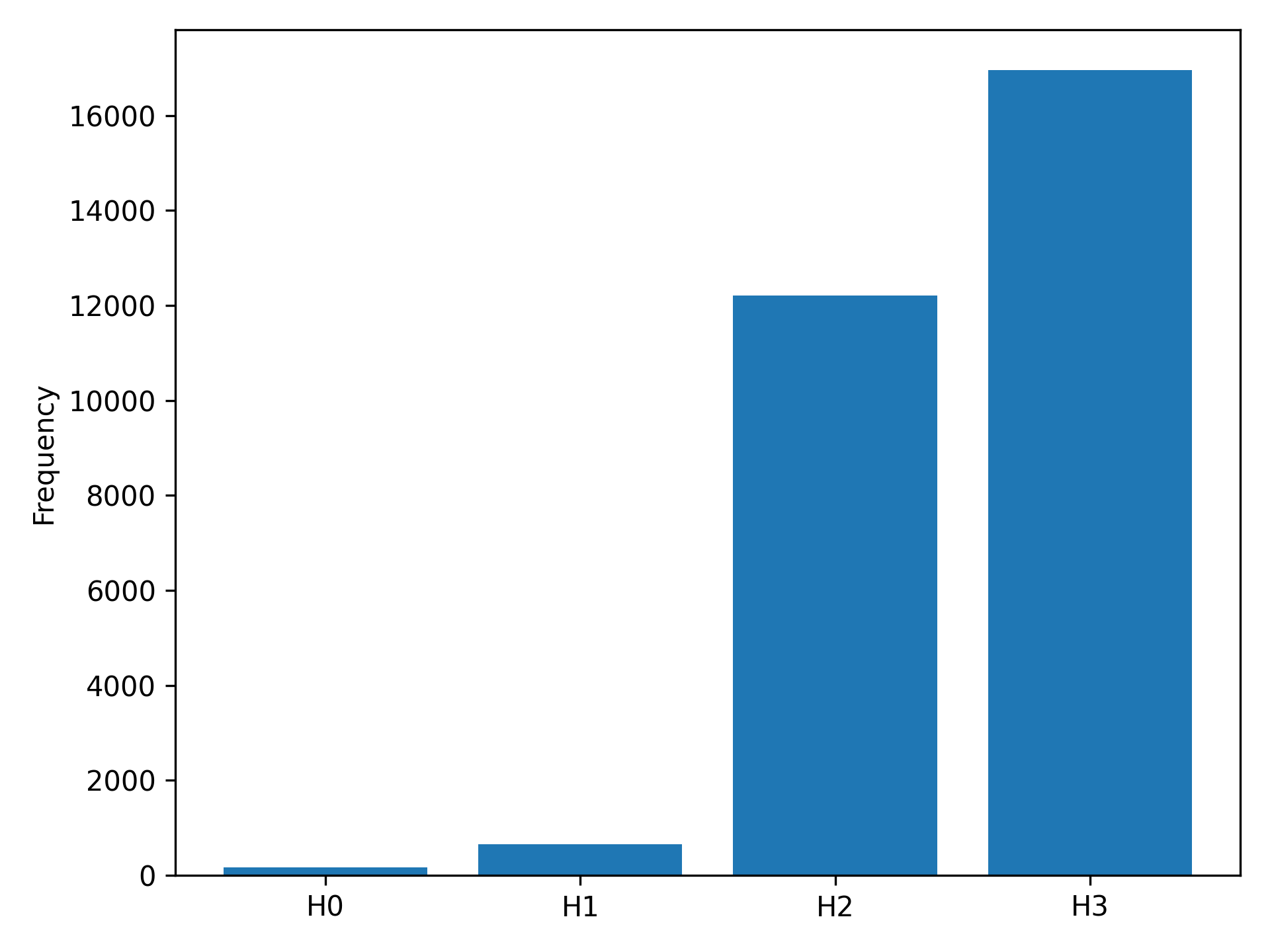}
        \caption{Top-1 hop selection count.}
    \end{subfigure}\hfill
    \begin{subfigure}{0.34\textwidth}
        \includegraphics[width=\linewidth]{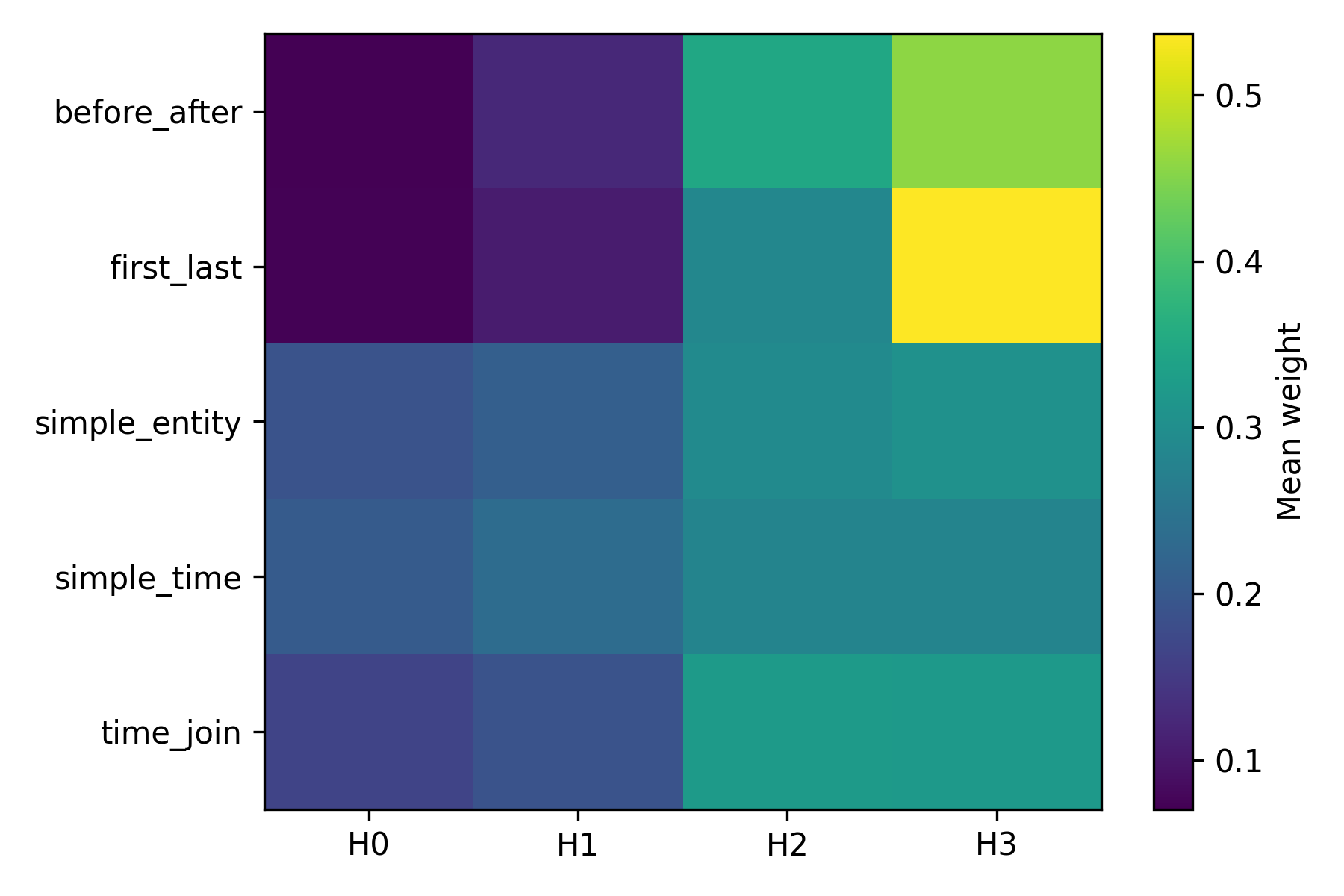}
        \caption{Hop weight heatmap by query type.}
    \end{subfigure}
    \caption{Behavioral dynamics of the differentiable working memory ($K=4$) across $30,000$ CronQuestions validation instances. The network dynamically routes computation depth, shifting attention mass to H3 for compositional temporal queries.}
    \label{fig:hop_analysis}
\end{figure*}

\paragraph{Hop Attention Analysis and Iterative Refinement.}
To better understand how SABET-QA exploits its multi-hop reasoning mechanism, we analyze the aggregation weights ($\beta_k$) and top-1 hop selections over $30,000$ validation queries from the CronQuestions dataset. As shown in Figure~\ref{fig:hop_analysis}(a,b), the model consistently favors deeper reasoning states: the mean aggregation weight increases monotonically from $0.137$ (Hop 0) to $0.400$ (Hop 3), while Hop 3 is selected as the dominant reasoning state in $56.5\%$ of the queries.

Further stratification by question type (Figure~\ref{fig:hop_analysis}(c)) reveals adaptive reasoning depth across different temporal reasoning tasks. Whereas relatively simple query types distribute attention more evenly across hops, compositionally complex queries such as \textit{first\_last} ($81.6\%$ Hop 3 selection) and \textit{before\_after} ($54.3\%$ Hop 3 selection) place the majority of their attention on the final hop. These results indicate that the differentiable working memory learns to postpone final predictions until sufficient multi-step temporal evidence has been accumulated, demonstrating its ability to adapt computation depth according to reasoning complexity.

\vspace{-1pt}
\paragraph{Ablation of architectural components.}
Table~\ref{tab:ablation} isolates the contribution of each design
choice in SABET-QA-Hard on Complex-CronQuestions.  Removing entity
contextualization (EC) drops Hits@1 by \textbf{10.0} points
(\textbf{0.807} $\rightarrow$ \textbf{0.707}), showing that grounding
slot representations in question text is essential when relations are
lexically ambiguous.  Removing bidirectional entity--time scoring
(BETS) causes the largest single-component drop
(\textbf{--14.8} points, to \textbf{0.659}), confirming that forward
and backward scorers capture complementary directional biases;
neither direction alone suffices for head--tail disambiguation.
Removing hard supervision (HS) degrades performance to the non-hard
variant level (\textbf{0.524}), demonstrating that temporal hint
injection provides orthogonal signal when interval boundaries are
available.  Crucially, no partial configuration approaches the full
model; the gain over the best ablated variant is \textbf{18.3}
Hits@1 points, establishing that slot contextualization, bidirectional
scoring, and hard supervision are \emph{jointly} necessary rather than
redundant.

\paragraph{Which pretrained modules should be fine-tuned?}
Table~\ref{tab:ablation_freeze} examines whether updating the
pretrained language model (LM) and temporal knowledge embeddings
(TKE) during QA training improves performance.

\vspace{-1pt}
Under \emph{no hard supervision} (SABET-QA, top block), the best
Complex-CronQuestions result (\textbf{0.547} Hits@1) is obtained by
\emph{unfreezing the LM while keeping TKE frozen} outperforming
the fully frozen baseline (\textbf{0.524}) by \textbf{2.3} points.
Unfreezing TKE alone (\textbf{0.446}) or both modules
(\textbf{0.476}) underperforms the frozen baseline, suggesting that
updating TKE without hard boundaries introduces noise into an
already well-structured embedding space.

Under \emph{hard supervision} (SABET-QA-Hard, bottom block), the
fully frozen configuration already achieves \textbf{0.807} Hits@1.
Unfreezing the LM alone yields \textbf{0.803} (\textbf{--0.4}), a
difference that is likely not significant, while unfreezing TKE
alone (\textbf{0.759}) or both (\textbf{0.769}) degrades
performance.  The same ordering holds on CronQuestions: frozen
$\geq$ LM-unfrozen $>$ both-unfrozen $>$ TKE-unfrozen.

These results indicate that the pretrained TComplEx embeddings are
already well-optimized for temporal link prediction, and updating
them during QA fine-tuning hurts generalization.  The bottleneck is
instead the \emph{alignment} between the LM's question encoding and
the frozen TKG scoring space.  With hard supervision, this
alignment is already strong enough that unfreezing the LM provides
no benefit; under weak supervision, modest gains from an unfrozen
LM are possible, but they are quickly overshadowed once hard
temporal boundaries are available.

\section{Conclusion}
\label{sec:conclusion}



In this work, we presented SABET-QA, an iterative multi-hop framework for Temporal Knowledge Graph Question Answering. By introducing bidirectional entity and time scoring, our model mitigates head-tail directional ambiguity, a common failure mode in prior methods. A differentiable working memory progressively refines the latent reasoning state across hops, moving beyond single-pass inference, while slot-aware contextualization keeps entity and temporal representations grounded in question semantics throughout.

Experiments on CronQuestions, Complex-CronQuestions, MultiTQ, and TimeQuestions show consistent improvements over strong baselines, with the largest gains on complex questions. This validates our hypothesis that iterative refinement and bidirectional temporal scoring are critical for multi-step temporal reasoning.

\newpage

\section*{Limitations}
\label{sec:limitations}

Despite these results, SABET-QA has several limitations. First, the approach relies on pretrained temporal KG embeddings and frozen language-model/KG components during QA training, which may constrain adaptability to new domains or evolving graph structures. Third, the method assumes access to entity and timestamp grounding when available, so performance drops when such annotations are missing or unreliable. When gold entity mentions are unavailable, SABET-QA relies on downstream NER systems (e.g., Flair \cite{akbik2019flair} on MultiTQ). Error propagation from noisy slot extractions inherently introduces a performance degradation compared to gold-annotated mentions, highlighting a dependency on upstream entity linking accuracy in end-to-end setups. Finally, the iterative multi-hop design adds architectural complexity and may increase computational cost compared with simpler single-pass models. These points are consistent with the paper’s training setup and the way the model is grounded in pretrained TComplEx-based representations.

\section*{Ethical Considerations}
This work is based on structured benchmark data and is intended for research on temporal reasoning. It is also important to note that the model may inherit biases, gaps, or factual errors from the underlying knowledge graphs and pretrained embeddings, and such issues can affect prediction quality. In addition, because the model is evaluated on benchmark datasets rather than sensitive personal data, the main ethical concerns relate to reproducibility, disclosure, and responsible interpretation of results rather than privacy.


\clearpage
\bibliography{custom,biblio}

\onecolumn
\newpage
\appendix

\section{Temporal KG Embedding Model}
\label{app:tcomplex}



To train the temporal knowledge-graph embeddings used in our QA models, we adopt \textit{TComplEx}~\cite{lacroix2020tensordecompositionstemporalknowledge}, following a broader family of complex-valued embedding models for knowledge graphs. In particular, ComplEx~\cite{trouillon2016complexembeddingssimplelink} represents entities and relations as complex vectors and scores a fact via the real part of a multilinear product, while TComplEx extends this formulation by introducing a timestamp embedding for temporal facts. Related temporal variants such as TNTComplEx and TimePlex further enrich this framework with additional time-aware parameterizations of relations and temporal representations~\cite{lacroix2020tensordecompositionstemporalknowledge,jain-etal-2020-temporal}. 

We use TComplEx as the default KG embedding backend for all QA models in order to ensure a controlled comparison across methods, since our focus is on evaluating the QA architecture rather than the choice of temporal KGE model. This design choice is therefore not essential to the proposed QA framework and can be replaced by another temporal embedding method without changing the core model. Nevertheless, TComplEx is a natural and widely used choice in the literature, which makes it suitable for our experiments~\cite{Ruffinelli2020You}.

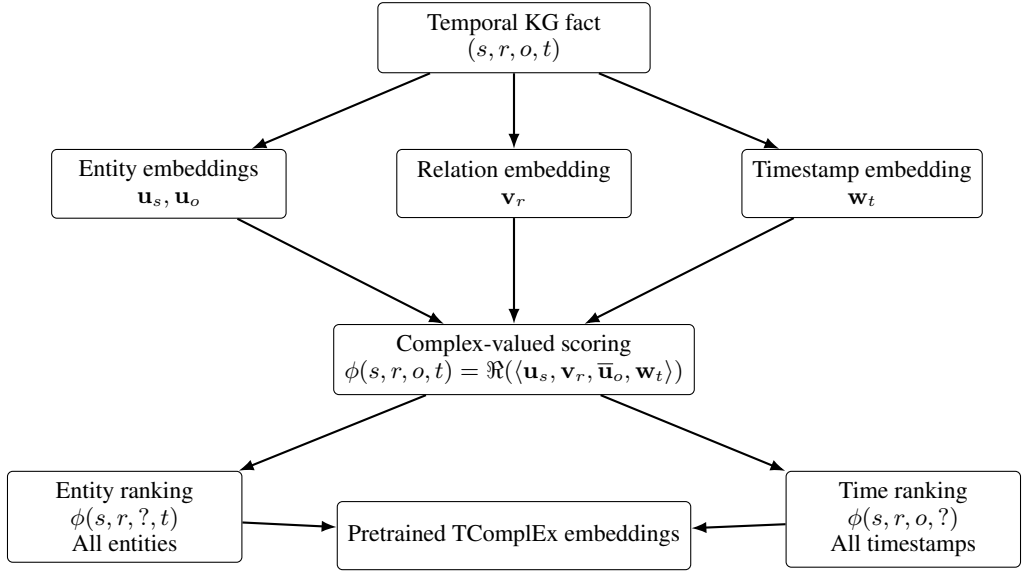
\begin{figure*}[t]
\centering
\begin{tikzpicture}[
    node distance = 10mm and 12mm,
    font=\small,
    box/.style = {
        draw,
        rounded corners=2pt,
        align=center,
        minimum height=9mm,
        minimum width=36mm,
        inner sep=4pt
    },
    smallbox/.style = {
        draw,
        rounded corners=2pt,
        align=center,
        minimum height=9mm,
        minimum width=31mm,
        inner sep=4pt
    },
    arrow/.style = {-{Latex[length=2mm]}, thick}
]

\node[box] (input) {Temporal KG fact\\$(s,r,o,t)$};

\node[smallbox, below left=of input] (ent) {Entity embeddings\\$\mathbf{u}_s,\mathbf{u}_o$};
\node[smallbox, below=of input] (rel) {Relation embedding\\$\mathbf{v}_r$};
\node[smallbox, below right=of input] (time) {Timestamp embedding\\$\mathbf{w}_t$};

\node[box, below=14mm of rel] (comp) {Complex-valued scoring\\
$\phi(s,r,o,t)=\Re\!\left(\langle \mathbf{u}_s,\mathbf{v}_r,\overline{\mathbf{u}}_o,\mathbf{w}_t\rangle\right)$};

\node[smallbox, below left=of comp] (entpred) {Entity ranking\\$\phi(s,r,?,t)$\\All entities};
\node[smallbox, below right=of comp] (timepred) {Time ranking\\$\phi(s,r,o,?)$\\All timestamps};

\node[box, below=14mm of comp] (model) {Pretrained TComplEx embeddings};

\draw[arrow] (input) -- (ent);
\draw[arrow] (input) -- (rel);
\draw[arrow] (input) -- (time);

\draw[arrow] (ent) -- (comp);
\draw[arrow] (rel) -- (comp);
\draw[arrow] (time) -- (comp);

\draw[arrow] (comp) -- (entpred);
\draw[arrow] (comp) -- (timepred);

\draw[arrow] (entpred) -- (model);
\draw[arrow] (timepred) -- (model);

\end{tikzpicture}
\caption{TComplEx pipeline used to initialize the temporal KG embeddings. The model learns complex-valued embeddings for entities, relations, and timestamps, and supports both entity ranking and timestamp ranking through the same compositional scoring function.}
\label{fig:temp_kge_pipeline}
\end{figure*}

Figure~\ref{fig:temp_kge_pipeline} summarizes the TComplEx pipeline. The model maps a temporal fact to entity, relation, and timestamp embeddings, combines them through complex-valued interactions, and supports both entity prediction and timestamp prediction.

For the MultiTQ dataset, temporal answers can be expressed at different granularities, including specific days, months, or years. To accommodate this variability, we trained the temporal knowledge graph embeddings on an augmented version of the original TKG. Specifically, for each temporal fact, we added additional facts corresponding to coarser temporal resolutions by converting timestamps into their associated month- and year-level representations. For example, a fact associated with a specific day was duplicated with equivalent month-level and year-level timestamps.

\section{SABET-QA Details}
\label{app:imsabt}

\begin{figure*}[t]
\centering
\resizebox{\textwidth}{!}{
\begin{tikzpicture}[
    node distance = 10mm and 12mm,
    font=\small,
    box/.style = {
        draw,
        rounded corners=2pt,
        align=center,
        minimum height=9mm,
        minimum width=38mm,
        inner sep=4pt
    },
    widebox/.style = {
        draw,
        rounded corners=2pt,
        align=center,
        minimum height=9mm,
        minimum width=52mm,
        inner sep=4pt
    },
    smallbox/.style = {
        draw,
        rounded corners=2pt,
        align=center,
        minimum height=8mm,
        minimum width=28mm,
        inner sep=3pt
    },
    arrow/.style = {-{Latex[length=2mm]}, thick}
]

\node[widebox] (q) {Question tokens\\$\mathbf{x},\mathbf{m}$};

\node[box, below=of q] (lm) {Pretrained LM};

\node[box, below=of lm] (proj) {Text projection\\to TKG space ($D$)};

\node[widebox, right=46mm of q] (slots)
{Head/tail/time\\slot embeddings};

\node[box, below=of slots] (lookup)
{Shared TKG\\embedding lookup};

\node[widebox, below=20mm of proj] (attn)
{Cross-attention slot contextualization};

\node[box, below=of attn] (gate)
{Gated residual fusion};

\node[smallbox, below left=9mm and 8mm of gate] (head)
{$\hat{\mathbf{e}}_h$};

\node[smallbox, below=9mm of gate] (tail)
{$\hat{\mathbf{e}}_t$};

\node[smallbox, below right=9mm and 8mm of gate] (time)
{$\hat{\mathbf{e}}_\tau$};

\draw[arrow] (q) -- (lm);
\draw[arrow] (lm) -- (proj);

\draw[arrow] (slots) -- (lookup);

\draw[arrow] (proj) -- (attn);
\draw[arrow] (lookup) |- (attn);

\draw[arrow] (attn) -- (gate);

\draw[arrow] (gate) -- (head);
\draw[arrow] (gate) -- (tail);
\draw[arrow] (gate) -- (time);

\end{tikzpicture}
}
\caption{
Question encoding and slot contextualization in SABER-TQA.
Question tokens are encoded by a pretrained language model and projected into the temporal KG embedding space. Head, tail, and temporal slot embeddings are retrieved from the shared TKG table and attend to the question representation, followed by gated residual fusion to produce contextualized role-specific vectors.
}
\label{fig:saberqa_text_slots}
\end{figure*}
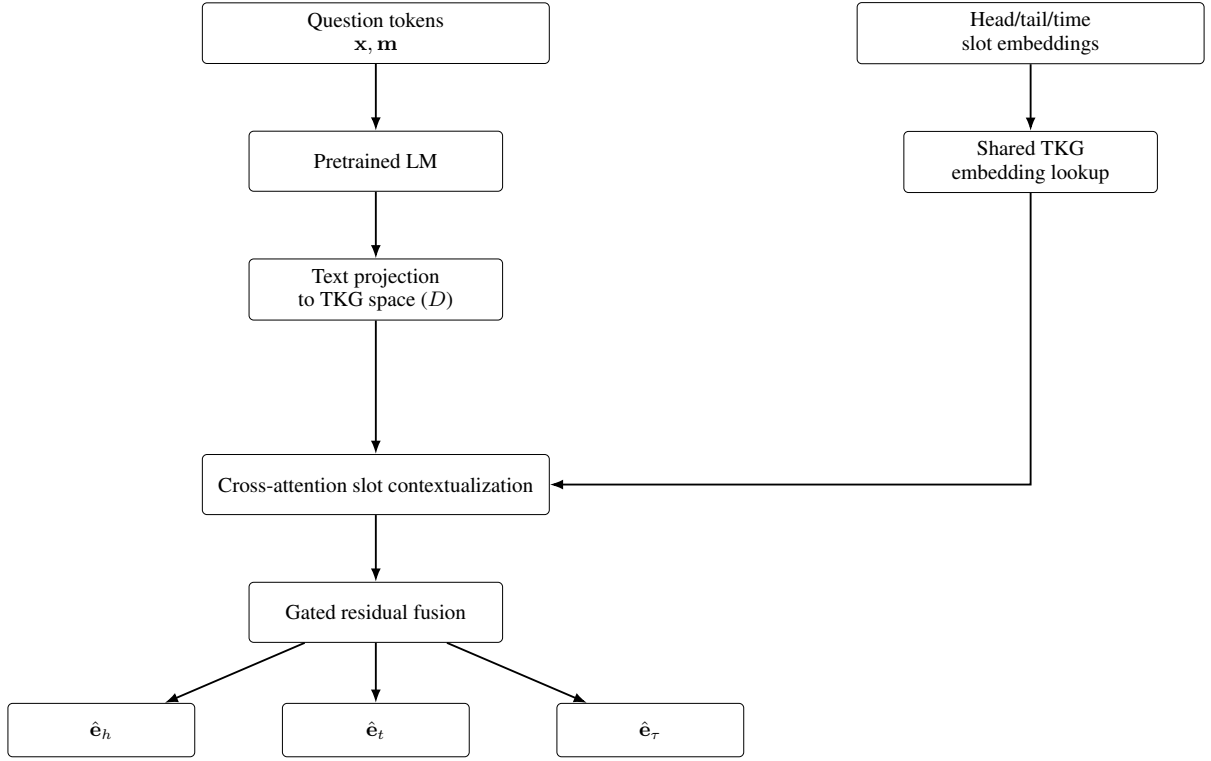

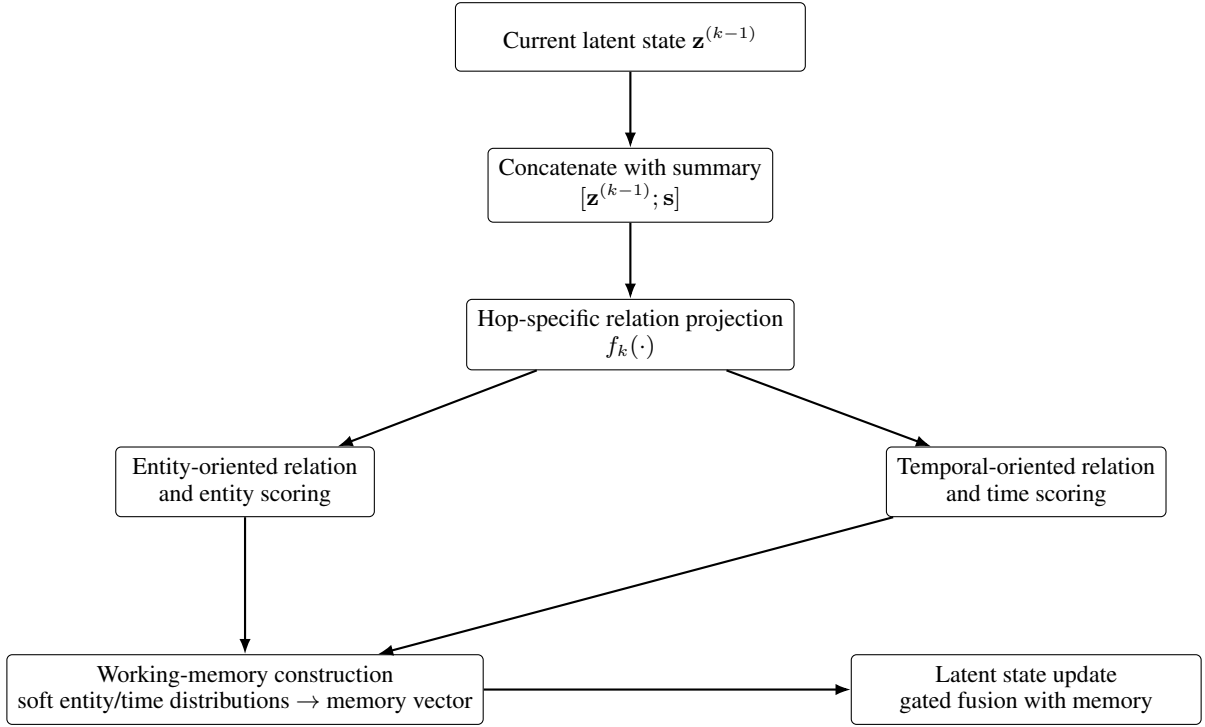
\begin{figure*}[t]
\centering
\resizebox{\textwidth}{!}{
\begin{tikzpicture}[
    node distance = 10mm and 12mm,
    font=\small,
    box/.style = {
        draw,
        rounded corners=2pt,
        align=center,
        minimum height=9mm,
        minimum width=34mm,
        inner sep=4pt
    },
    widebox/.style = {
        draw,
        rounded corners=2pt,
        align=center,
        minimum height=9mm,
        minimum width=46mm,
        inner sep=4pt
    },
    arrow/.style = {-{Latex[length=2mm]}, thick}
]

\node[widebox] (state) {Current latent state $\mathbf{z}^{(k-1)}$};

\node[box, below=of state] (concat)
{Concatenate with summary\\$[\mathbf{z}^{(k-1)};\mathbf{s}]$};

\node[box, below=of concat] (proj)
{Hop-specific relation projection\\$f_k(\cdot)$};

\node[box, below left=of proj] (ent)
{Entity-oriented relation\\and entity scoring};

\node[box, below right=of proj] (time)
{Temporal-oriented relation\\and time scoring};

\node[widebox, below=18mm of ent] (mem)
{Working-memory construction\\soft entity/time distributions $\rightarrow$ memory vector};

\node[widebox, below=18mm of time] (next)
{Latent state update\\gated fusion with memory};

\draw[arrow] (state) -- (concat);
\draw[arrow] (concat) -- (proj);

\draw[arrow] (proj) -- (ent);
\draw[arrow] (proj) -- (time);

\draw[arrow] (ent) -- (mem);
\draw[arrow] (time) -- (mem);

\draw[arrow] (mem) -- (next);

\end{tikzpicture}
}
\caption{
Iterative hop-wise reasoning in SABET-QA.
At each hop, the current latent reasoning state is combined with the global question summary and transformed into a hop-specific relation representation. Dedicated entity-oriented and temporal-oriented projections generate entity and timestamp scores, whose soft distributions are converted into a working-memory representation used to update the latent state for the next reasoning hop.
}
\label{fig:saberqa_hops}
\end{figure*}

This appendix provides the complete mathematical specification and pseudocode for SABET-QA, matching the implementation provided in the anonymized supplementary material.

\subsection{Question Encoding and Slot Contextualization}
\label{app:enc}

Given tokenized question $\mathbf{x}=(x_1,\dots,x_L)$ and attention mask $\mathbf{m}$, the pretrained LM produces hidden states $\mathbf{H}=\mathrm{LM}(\mathbf{x},\mathbf{m})\in\mathbb{R}^{L\times 768}$. These are projected into the TKG space by a feed-forward network $f_{\text{text}}$ (linear $\to$ LayerNorm $\to$ GELU $\to$ dropout):
\begin{equation}
\mathbf{T}=f_{\text{text}}(\mathbf{H})\in\mathbb{R}^{L\times D}.
\end{equation}

Head, tail, and timestamp slot embeddings $\mathbf{e}_h,\mathbf{e}_t,\mathbf{e}_\tau$ are retrieved from the shared entity/time embedding table. They are stacked as queries for multi-head cross-attention over the projected question tokens:
\begin{equation}
\mathbf{C}=\mathrm{Attn}\big([\mathbf{e}_h;\mathbf{e}_t;\mathbf{e}_\tau],\mathbf{T},\mathbf{T}\big),
\end{equation}
with key padding derived from $\mathbf{m}$. The outputs $\mathbf{c}_h,\mathbf{c}_t,\mathbf{c}_\tau$ are layer-normalized and fused with gated residuals:
\begin{equation}
g_h=\sigma\big(W_g[\mathbf{e}_h;\mathbf{c}_h]\big),\qquad
\hat{\mathbf{e}}_h=g_h\odot\mathbf{e}_h+(1-g_h)\odot\mathbf{c}_h,
\end{equation}
and analogously for $\hat{\mathbf{e}}_t$ and $\hat{\mathbf{e}}_\tau$.

\subsection{Global Summary and Base Relation}
\label{app:summary}

A global summary vector is built from the projected token sequence using the concatenation of the [CLS] representation, masked mean pooling, and masked max pooling:
\begin{equation}
\mathbf{s}=f_{\text{sum}}\big([\mathbf{T}_{\text{CLS}};\mathrm{Mean}(\mathbf{T});\mathrm{Max}(\mathbf{T})]\big)\in\mathbb{R}^{D},
\end{equation}
where $f_{\text{sum}}$ is a learned projection. The base latent state is then constructed from the summary and the contextualized slots:
\begin{equation}
\mathbf{z}^{(0)}=f_{\text{rel}}\big([\mathbf{s};\hat{\mathbf{e}}_h;\hat{\mathbf{e}}_t;\hat{\mathbf{e}}_\tau]\big)\in\mathbb{R}^{D},
\end{equation}
where $f_{\text{rel}}$ is a two-layer MLP ($4D\to 2D\to D$) with GELU and dropout.

\subsection{Hop-Specific Relation Projection}
\label{app:hop-proj}

At hop $k$, the current state and the global summary are concatenated and projected by a hop-specific MLP $f_k$:
\begin{equation}
\mathbf{r}^{(k)}=f_k\big([\mathbf{z}^{(k-1)};\mathbf{s}]\big),
\end{equation}
where $f_k$ maps $2D\to 2D\to D$ with LayerNorm, GELU, and dropout.

This relation is then factorized into an entity-oriented view and a time-oriented view using dedicated projection heads:
\begin{equation}
\mathbf{r}_{\text{ent}}^{(k)}=g_{\text{ent}}(\mathbf{r}^{(k)}),\qquad
\mathbf{r}_{\text{time}}^{(k)}=g_{\text{time}}(\mathbf{r}^{(k)}),
\end{equation}
with each head implemented as a small MLP over $D$-dimensional inputs.

\subsection{Optional Temporal Hint Injection}
\label{app:hint}

When auxiliary temporal endpoints $(t_1,t_2)$ are available, their TKG embeddings are used as a hard temporal hint. Let $\mathbf{e}_{t_1}$ and $\mathbf{e}_{t_2}$ denote the corresponding timestamp embeddings. A learned gate controls how much of this hint is injected into the current latent state:
\begin{equation}
\boldsymbol{\tau}^{(k)}=\sigma\big(W_{\tau}[\mathbf{z}^{(k-1)};\mathbf{e}_{t_1};\mathbf{e}_{t_2}]\big),
\end{equation}
and the state is updated as
\begin{equation}
\mathbf{z}^{(k-1)}\leftarrow \mathbf{z}^{(k-1)}+\boldsymbol{\tau}^{(k)}\odot(\mathbf{e}_{t_1}+\mathbf{e}_{t_2}).
\end{equation}
This hint is optional and only contributes when the corresponding timestamps are available.

\subsection{Bidirectional Entity Scoring}
\label{app:bi-ent}

Using the contextualized slots $\hat{\mathbf{e}}_h,\hat{\mathbf{e}}_t,\hat{\mathbf{e}}_\tau$, the forward and backward TComplEx scores are:
\begin{align}
\mathbf{s}_{\text{ent}}^{\rightarrow(k)} &= \mathrm{Score}_{\text{TComplEx}}(\hat{\mathbf{e}}_h,\hat{\mathbf{e}}_t,\mathbf{r}_{\text{ent}}^{(k)},\hat{\mathbf{e}}_\tau), \\
\mathbf{s}_{\text{ent}}^{\leftarrow(k)} &= \mathrm{Score}_{\text{TComplEx}}(\hat{\mathbf{e}}_t,\hat{\mathbf{e}}_h,\mathbf{r}_{\text{ent}}^{(k)},\hat{\mathbf{e}}_\tau).
\end{align}

They are fused by a learned scalar gate conditioned on the global summary:
\begin{equation}
\alpha_{\text{ent}}^{(k)}=\sigma(W_{\text{ent}}\mathbf{s}),\qquad
\mathbf{s}_{\text{ent}}^{(k)}=\alpha_{\text{ent}}^{(k)}\mathbf{s}_{\text{ent}}^{\rightarrow(k)}+\big(1-\alpha_{\text{ent}}^{(k)}\big)\mathbf{s}_{\text{ent}}^{\leftarrow(k)}.
\end{equation}

\subsection{Bidirectional Time Scoring}
\label{app:bi-time}

Temporal candidates are scored analogously:
\begin{align}
\mathbf{s}_{\text{time}}^{\rightarrow(k)} &= \mathrm{Score}_{\text{time}}(\hat{\mathbf{e}}_h,\hat{\mathbf{e}}_t,\mathbf{r}_{\text{time}}^{(k)}), \\
\mathbf{s}_{\text{time}}^{\leftarrow(k)} &= \mathrm{Score}_{\text{time}}(\hat{\mathbf{e}}_t,\hat{\mathbf{e}}_h,\mathbf{r}_{\text{time}}^{(k)}).
\end{align}
The two score vectors are fused by a gate conditioned on the global summary:
\begin{equation}
\alpha_{\text{time}}^{(k)}=\sigma(W_{\text{time}}\mathbf{s}),\qquad
\mathbf{s}_{\text{time}}^{(k)}=\alpha_{\text{time}}^{(k)}\mathbf{s}_{\text{time}}^{\rightarrow(k)}+\big(1-\alpha_{\text{time}}^{(k)}\big)\mathbf{s}_{\text{time}}^{\leftarrow(k)}.
\end{equation}

\subsection{Working-Memory Update}
\label{app:mem}

After scoring, the model converts the entity and time scores into soft distributions:
\begin{equation}
\mathbf{p}_{\text{ent}}^{(k)}=\mathrm{softmax}(\mathbf{s}_{\text{ent}}^{(k)}),\qquad
\mathbf{p}_{\text{time}}^{(k)}=\mathrm{softmax}(\mathbf{s}_{\text{time}}^{(k)}).
\end{equation}

These are used to compute expected entity and time embeddings:
\begin{equation}
\bar{\mathbf{e}}_{\text{ent}}^{(k)}=\mathbf{p}_{\text{ent}}^{(k)}\mathbf{E}_{\text{ent}},\qquad
\bar{\mathbf{e}}_{\text{time}}^{(k)}=\mathbf{p}_{\text{time}}^{(k)}\mathbf{E}_{\text{time}},
\end{equation}
where $\mathbf{E}_{\text{ent}}$ and $\mathbf{E}_{\text{time}}$ are the entity and timestamp embedding tables from the TKBC model.

The two expected embeddings are summed, projected into a memory vector, and used to refine the latent state through an attention-based gated update:
\begin{equation}
\mathbf{m}^{(k)}=f_{\text{mem}}\big(\bar{\mathbf{e}}_{\text{ent}}^{(k)}+\bar{\mathbf{e}}_{\text{time}}^{(k)}\big),
\end{equation}
followed by
\begin{equation}
\mathbf{u}^{(k)}=\mathrm{Attn}\big(\mathbf{z}^{(k-1)}\uparrow,\mathbf{m}^{(k)}\uparrow,\mathbf{m}^{(k)}\uparrow\big)\downarrow,
\end{equation}
and a gated residual fusion:
\begin{align}
\gamma^{(k)} &= \sigma\big(W_{\gamma}[\mathbf{z}^{(k-1)};\mathbf{u}^{(k)}]\big),\\
\mathbf{z}^{(k)} &= \gamma^{(k)}\odot\mathbf{z}^{(k-1)}+(1-\gamma^{(k)})\odot\mathbf{u}^{(k)}.
\end{align}

This mechanism lets the model carry forward predictions from earlier hops and progressively refine its hypothesis.

\subsection{Hop Aggregation and Training Objective}
\label{app:agg}

After $K$ hops, a hop-selection distribution is computed from the global summary:
\begin{equation}
\boldsymbol{\beta}=\mathrm{softmax}(W_{\text{hop}}\mathbf{s}).
\end{equation}

The final predictions are weighted sums of the hop-specific scores:
\begin{equation}
\mathbf{s}_{\text{ent}}=\sum_{k=1}^{K}\beta_k\mathbf{s}_{\text{ent}}^{(k)},\qquad
\mathbf{s}_{\text{time}}=\sum_{k=1}^{K}\beta_k\mathbf{s}_{\text{time}}^{(k)}.
\end{equation}

The final output is the concatenation $[\mathbf{s}_{\text{ent}};\mathbf{s}_{\text{time}}]$, trained with cross-entropy against the gold answer distribution. The TKBC embeddings used for scoring can be kept fixed during QA training when the frozen setting is enabled.

\subsection{Pseudocode}
\label{app:pseudo}

Algorithm~\ref{alg:saberqa} gives a compact PyTorch-style pseudocode matching the implementation.

\begin{algorithm}[H]
\footnotesize
\caption{SABET-QA forward pass.}
\label{alg:saberqa}
\begin{algorithmic}[1]
\Require question tokens $\mathbf{x}$, mask $\mathbf{m}$, heads, tails, times, optional $(t_1,t_2)$
\Ensure concatenated entity/time scores

\State $\mathbf{H} \gets \mathrm{LM}(\mathbf{x}, \mathbf{m})$
\State $\mathbf{T} \gets \text{text\_proj}(\mathbf{H})$
\State $\mathbf{e}_h, \mathbf{e}_t, \mathbf{e}_\tau \gets \text{lookup}(\text{heads}, \text{tails}, \text{times})$
\State $\hat{\mathbf{e}}_h, \hat{\mathbf{e}}_t, \hat{\mathbf{e}}_\tau \gets \text{slot\_context}(\mathbf{T}, \mathbf{m}, \mathbf{e}_h, \mathbf{e}_t, \mathbf{e}_\tau)$
\State $\mathbf{s} \gets \text{summary}(\mathbf{T}, \mathbf{m})$
\State $\mathbf{z}^{(0)} \gets \text{rel\_head}([\mathbf{s};\hat{\mathbf{e}}_h;\hat{\mathbf{e}}_t;\hat{\mathbf{e}}_\tau])$

\For{$k = 1$ to $K$}
    \If{$t_1,t_2$ are available}
        \State $\mathbf{z}^{(k-1)} \gets \text{inject\_temporal\_hint}(\mathbf{z}^{(k-1)}, t_1, t_2)$
    \EndIf
    \State $\mathbf{r}^{(k)} \gets \text{hop\_proj}_k([\mathbf{z}^{(k-1)};\mathbf{s}])$
    \State $\mathbf{r}_{\text{ent}}^{(k)} \gets \text{ent\_proj}(\mathbf{r}^{(k)})$
    \State $\mathbf{r}_{\text{time}}^{(k)} \gets \text{time\_proj}(\mathbf{r}^{(k)})$
    \State $\mathbf{s}_{\text{ent}}^{(k)} \gets \text{bidirectional\_entity\_score}(\hat{\mathbf{e}}_h,\hat{\mathbf{e}}_t,\hat{\mathbf{e}}_\tau,\mathbf{r}_{\text{ent}}^{(k)},\mathbf{s})$
    \State $\mathbf{s}_{\text{time}}^{(k)} \gets \text{bidirectional\_time\_score}(\hat{\mathbf{e}}_h,\hat{\mathbf{e}}_t,\mathbf{r}_{\text{time}}^{(k)},\mathbf{s})$
    \If{$k < K$}
        \State $\mathbf{z}^{(k)} \gets \text{memory\_update}(\mathbf{z}^{(k-1)}, \mathbf{s}_{\text{ent}}^{(k)}, \mathbf{s}_{\text{time}}^{(k)})$
    \EndIf
\EndFor

\State $\boldsymbol{\beta} \gets \mathrm{Softmax}(W_{\text{hop}}\mathbf{s})$
\State $\mathbf{s}_{\text{ent}} \gets \sum_{k=1}^{K} \beta_k\,\mathbf{s}_{\text{ent}}^{(k)}$
\State $\mathbf{s}_{\text{time}} \gets \sum_{k=1}^{K} \beta_k\,\mathbf{s}_{\text{time}}^{(k)}$
\State \Return $[\mathbf{s}_{\text{ent}};\mathbf{s}_{\text{time}}]$
\end{algorithmic}
\end{algorithm}

\section{Dataset Statistics}
\label{app:datasets}
\begin{table}[H]
\centering
\small
\begin{tabular}{lrrrrl}
\toprule
Dataset & Train & Valid & Test & Total & Split Ratio (Tr/Va/Te) \\
\midrule
CronQuestions & 350,000 & 30,000 & 30,522 & 410,522 & 85.3\% / 7.3\% / 7.4\% \\
Complex-CronQuestions & 35,795 & 5,020 & 5,528 & 46,343 & 77.2\% / 10.8\% / 11.9\% \\
MultiTQ & 386,787 & 57,979 & 54,584 & 499,350 & 77.5\% / 11.6\% / 10.9\% \\
TimeQuestions & 6,970 & 3,236 & 3,237 & 13,443 & 51.8\% / 24.1\% / 24.1\% \\
\bottomrule
\end{tabular}
\caption{Statistics of the temporal question answering datasets used in our experiments.}
\label{tab:dataset_meta}
\end{table}

\begin{table}[H]
\centering
\small
\begin{tabular}{lccrrrrr}
\toprule
Dataset & Entities & Relations & Timestamps & Train Facts & Valid Facts & Test Facts & Total Facts \\
\midrule
CronQuestions & 125,726 & 406 & 9621 & 323,635 & 5000 & 5000 & 333,635 \\
Complex-CronQuestions & 125,726 & 406 & 9621 & 323,635 & 5000 & 5000 & 333,635 \\
MultiTQ & 10,488 & 502 & 4,017 & 322,958 & 69,224 & 69,147 & 461,329 \\
TimeQuestions & 118,010 & 884 & 1,636 & 227,564 & 4,997 & 4,998 & 240,597 \\
\bottomrule
\end{tabular}
\caption{Statistics of the temporal knowledge graphs associated with each benchmark.}
\label{tab:tkg_meta}
\end{table}

\begin{table*}[t]
\centering
\small

\begin{subtable}[t]{0.48\textwidth}
\centering
\begin{tabular}{lrrr}
\toprule
\multicolumn{4}{c}{CronQuestions} \\
\midrule
Answer Type & Train & Valid & Test \\
\midrule
Entity & 225,672 & 19,362 & 19,524 \\
Time & 124,328 & 10,638 & 10,476 \\
\midrule
Total & 350,000 & 30,000 & 30,000 \\
\bottomrule
\end{tabular}
\caption{Answer types in CronQuestions.}
\label{tab:cron_answer_types}
\end{subtable}
\hfill
\begin{subtable}[t]{0.48\textwidth}
\centering
\begin{tabular}{lrrr}
\toprule
\multicolumn{4}{c}{Complex-CronQuestions} \\
\midrule
Answer Type & Train & Valid & Test \\
\midrule
Entity & 23,029 & 3,340 & 3,382 \\
Time & 12,766 & 1,680 & 1,624 \\
\midrule
Total & 35,795 & 5,020 & 5,006 \\
\bottomrule
\end{tabular}
\caption{Answer types in Complex-CronQuestions.}
\label{tab:complex_answer_types}
\end{subtable}

\vspace{0.75em}

\begin{subtable}[t]{0.48\textwidth}
\centering
\begin{tabular}{lrrr}
\toprule
\multicolumn{4}{c}{MultiTQ} \\
\midrule
Answer Type & Train & Valid & Test \\
\midrule
Entity & 267,155 & 40,565 & 38,700 \\
Time & 119,632 & 17,414 & 15,884 \\
\midrule
Total & 386,787 & 57,979 & 54,584 \\
\bottomrule
\end{tabular}
\caption{Answer types in MultiTQ.}
\label{tab:multitq_answer_types}
\end{subtable}
\hfill
\begin{subtable}[t]{0.48\textwidth}
\centering
\begin{tabular}{lrrr}
\toprule
\multicolumn{4}{c}{TimeQuestions} \\
\midrule
Answer Type & Train & Valid & Test \\
\midrule
Entity & 4,589 & 2,292 & 2,340 \\
Time & 2,381 & 944 & 897 \\
\midrule
Total & 6,970 & 3,236 & 3,237 \\
\bottomrule
\end{tabular}
\caption{Answer types in TimeQuestions.}
\label{tab:timequestions_answer_types}
\end{subtable}

\caption{Answer type distributions across the benchmarks.}
\label{tab:answer_type_distributions}
\end{table*}

We evaluate SABET-QA on four benchmark datasets for temporal question answering:
CronQuestions~\cite{saxena-etal-2021-question},
Complex-CronQuestions~\cite{CHEN2022109134},
MultiTQ~\cite{chen-etal-2023-multi}, and
TimeQuestions~\cite{Jia_2021}.

CronQuestions and Complex-CronQuestions are synthetic benchmarks derived from temporal Wikidata facts and focus on compositional temporal reasoning. MultiTQ contains large-scale automatically generated temporal questions involving diverse temporal operators and reasoning patterns. TimeQuestions consists of natural-language temporal questions collected from real-world sources and provides a complementary evaluation setting with realistic temporal information needs.

Table~\ref{tab:dataset_meta} summarizes the statistics of the question-answering datasets. Table~\ref{tab:tkg_meta} reports the characteristics of the associated temporal knowledge graphs. The datasets vary considerably in scale, ranging from 13K questions in TimeQuestions to nearly 500K questions in MultiTQ. Likewise, the underlying temporal knowledge graphs differ substantially in their numbers of entities, relations, timestamps, and temporal facts.

Table~\ref{tab:answer_type_distributions} further reports the distribution of answer types across the benchmarks. All datasets contain both entity-answer and timestamp-answer questions, providing a comprehensive evaluation of temporal reasoning capabilities across heterogeneous settings.

\section{Training Settings}
\label{app:training-details}

\begin{table}[t]
\centering
\small
\setlength{\tabcolsep}{5pt}
\begin{tabular}{ll}
\toprule
\textbf{Setting} & \textbf{Value} \\
\midrule
Optimizer & Adam~\cite{kingma2017adammethodstochasticoptimization} \\
Initial learning rate & $2 \times 10^{-4}$ (\,$6 \times 10^{-4}$ for Timequestions Dataset\,) \\
Maximum epochs & 20 (50 for Timequestions Dataset) \\
Training batch size & 150 \\
Validation batch size & 150 \\
Validation frequency & Every epoch \\
Learning-rate schedule & Linear warm-up + cosine annealing \\
Warm-up steps & $\min(200,\ 0.1 \times \text{total steps})$ \\
Warm-up start factor & 0.1 \\
Cosine minimum LR & $0.01 \times$ initial LR \\
Evaluation metrics & Hits@1, Hits@10 \\
Checkpoint selection & Best validation Hits@1 \\
Pretrained KG embeddings & Loaded from dataset-specific checkpoint \\
KG embedding update & Frozen during QA training (Except for Ablation Studies) \\
Language model update & Frozen during QA training (Except for Ablation Studies) \\
\bottomrule
\end{tabular}
\caption{Training settings used for all experiments unless otherwise noted.}
\label{tab:training_settings}
\end{table}

Table~\ref{tab:training_settings} summarizes the training configuration used in our experiments. All models were implemented in PyTorch~\cite{paszke2019pytorchimperativestylehighperformance}. The language model and pretrained KG embeddings were frozen during QA training.

\end{document}